\documentclass[conference]{IEEEtran}
\IEEEoverridecommandlockouts
\usepackage{cite}
\usepackage{amsmath,amssymb,amsfonts}
\usepackage{algorithm}
\usepackage{algpseudocode}
\usepackage{booktabs}
\usepackage{graphicx}
\usepackage{multirow}
\usepackage{textcomp}
\usepackage{xcolor}
\newcommand{\sd}[1]{$_{\pm #1}$}
\begin{document}
%%%%%%%%%%%%%%%%%%%%%%%%%%%%%%%%%%%%%%%%
\title{ReHear: Iterative Pseudo-Label Refinement for\\Semi-Supervised Speech Recognition via\\Audio Large Language Models}
%%%%%%%%%%%%%%%%%%%%%%%%%%%%%%%%%%%%%%%%
\author{%
\IEEEauthorblockN{Zefang Liu, Chenyang Zhu, Sangwoo Cho, Shi-Xiong Zhang}
\IEEEauthorblockA{\textit{Capital One}\\
USA\\
zefang.liu@capitalone.com}
}
%%%%%%%%%%%%%%%%%%%%%%%%%%%%%%%%%%%%%%%%
\maketitle
%%%%%%%%%%%%%%%%%%%%%%%%%%%%%%%%%%%%%%%%
\begin{abstract}
Semi-supervised learning in automatic speech recognition (ASR) typically relies on pseudo-labeling, which often suffers from confirmation bias and error accumulation due to noisy supervision. To address this limitation, we propose ReHear, a framework for iterative pseudo-label refinement that integrates an instruction-tuned, audio-aware large language model (LLM) into the self-training loop. Unlike conventional text-based correctors, our approach conditions the LLM on both the ASR hypothesis and the source audio, allowing it to recover phonetically accurate transcripts even from severe recognition errors. These refined pseudo-labels serve as high-fidelity targets for fine-tuning the ASR model in an iterative cycle. Experimental results across diverse benchmarks demonstrate that ReHear effectively mitigates error propagation and improves over both supervised and pseudo-labeling baselines.
\end{abstract}
%%%%%%%%%%%%%%%%%%%%%%%%%%%%%%%%%%%%%%%%
\begin{IEEEkeywords}
automatic speech recognition, large language models, semi-supervised learning, pseudo-labeling
\end{IEEEkeywords}
%%%%%%%%%%%%%%%%%%%%%%%%%%%%%%%%%%%%%%%%
\section{Introduction}

End-to-end (E2E) automatic speech recognition (ASR) systems~\cite{prabhavalkar2023end}, exemplified by models like Conformer~\cite{gulati2020conformer} and Whisper~\cite{radford2023robust}, have achieved state-of-the-art performance across a wide range of benchmarks. Although the utilization of massive, often weakly supervised web-scale datasets has endowed these models with impressive general capabilities, adapting them to the distinct acoustic and linguistic nuances of specific downstream tasks remains a significant challenge. Achieving high precision in specialized domains, distinct accents, or low-resource languages typically demands high-quality, human-annotated data. Consequently, the prohibitive cost and scarcity of such ground-truth labels create a significant bottleneck, severely limiting the effective deployment of these powerful models in scenarios where labeled data is sparse but unlabeled audio is abundant.

To mitigate the reliance on labeled data, semi-supervised learning (SSL), particularly through pseudo-labeling (PL), has emerged as a prevalent strategy. In this framework, a ``teacher'' ASR model generates transcriptions for a large corpus of unlabeled audio, which subsequently serve as targets to fine-tune a ``student'' model. However, the efficacy of this approach is fundamentally limited by error propagation. As the teacher model is prone to misrecognitions, training the student on these noisy pseudo-labels can reinforce errors and destabilize the optimization process. Although various filtering strategies based on confidence scores or uncertainty estimation have been proposed to curate the training data, they are often suboptimal; by discarding samples with low-confidence labels, these methods forfeit potentially valuable acoustic information contained in the discarded audio.

\begin{figure}[!t]
\centering
\includegraphics[width=\linewidth]{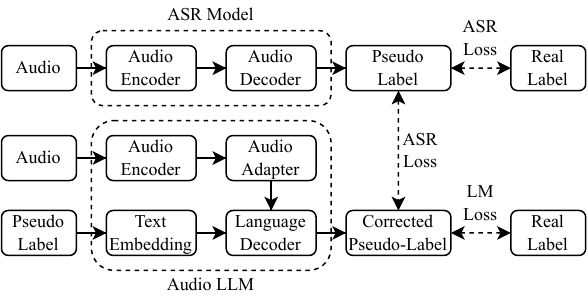}
\caption{Overview of the ReHear framework, where an audio-aware LLM refines ASR hypotheses via multimodal context to produce pseudo-labels that iteratively fine-tune the ASR model.}
\label{fig:framework}
\end{figure}

In this work, we diverge from the traditional filtering paradigm and instead propose to actively refine noisy pseudo-labels. We introduce \textbf{ReHear}, a novel semi-supervised ASR framework that leverages the emergent capabilities of multimodal audio large language models (Audio LLMs)~\cite{cui2025recent}. As illustrated in Figure~\ref{fig:framework}, our approach employs a multimodal corrector that conditions on both the initial ASR hypothesis and the raw source audio. By grounding corrections directly in the acoustic signal, the model can disambiguate phonetic uncertainties and rectify errors that are intractable for text-only correctors. Specifically, we establish a collaborative loop between a base ASR model and an Audio LLM, where the latter refines the ASR's initial predictions to generate high-fidelity targets for subsequent fine-tuning. This cycle fosters a self-improving dynamic, enabling the ASR system to effectively exploit large-scale unlabeled data while progressively mitigating error propagation. Crucially, ReHear distills the corrector's knowledge into the ASR model itself, so no LLM is required at inference time, leaving the deployed model's size and inference latency identical to the base recognizer, and offering a data- and deployment-efficient pathway to adapt ASR systems to new domains without extensive human annotation.

%%%%%%%%%%%%%%%%%%%%%%%%%%%%%%%%%%%%%%%%
\section{Related Work}

Pseudo-labeling (PL) has become a dominant paradigm for semi-supervised ASR~\cite{kahn2020self, chen2020semi}, significantly advanced by iterative pseudo-labeling (IPL) strategies where targets are continuously re-generated during training~\cite{xu2020iterative}. This process has been further refined through teacher-student frameworks utilizing exponential moving average (EMA)~\cite{manohar2021kaizen} or momentum-based updates~\cite{higuchi2021momentum}, alongside techniques for improved stability such as dynamic caching~\cite{likhomanenko2021slimipl} and convergent teacher initialization~\cite{tadevosyan2025unified}. To mitigate error propagation from noisy labels, standard approaches typically filter unlabeled data based on uncertainty metrics~\cite{khurana2021unsupervised, kim2025uncertainty}, confidence scores~\cite{jin2022filter}, or multi-model consensus~\cite{carofilis2025better}. Diverging from simple filtering, recent research attempts to actively correct noisy labels, utilizing methods ranging from modified Connectionist Temporal Classification (CTC) objectives~\cite{zhu2023alternative} and parameter-space bias correction~\cite{lin2026pseudo2real} to multimodal Speech LLMs that leverage source audio for correction~\cite{prakash2025better}. Whereas~\cite{prakash2025better} relies on multi-ASR fusion with a heavy Speech LLM applied at inference, ReHear uses a single audio-aware LLM corrector inside an iterative training loop and distills its knowledge into the ASR model, so no LLM is required at inference time. This design keeps the final deployed system lightweight while retaining the corrector's ability to repair phonetic errors from the source audio, at the cost of heavier offline training than confidence-filtered pseudo-labeling, a one-time expense that does not affect the LLM-free inference.

%%%%%%%%%%%%%%%%%%%%%%%%%%%%%%%%%%%%%%%%
\section{Methodology}

Our proposed \textbf{ReHear} framework orchestrates a self-improving training loop between a primary ASR model, denoted as $M_A$, and an audio-aware LLM functioning as a corrector, denoted as $M_L$. As outlined in Algorithm~\ref{alg:rehear}, the protocol iterates through four sequential stages: ASR inference, LLM training, LLM inference, and ASR training. This cyclic design fosters a symbiotic relationship where the two models mutually reinforce one another: an improved ASR model supplies more accurate initial hypotheses, which enables the corrector to generate higher-fidelity pseudo-labels for subsequent ASR optimization.

\begin{algorithm}[h]
\caption{ReHear: Iterative Pseudo-Label Refinement}
\label{alg:rehear}
\begin{algorithmic}[1]

\Statex \textbf{Models:} ASR model $M_A$, Audio LLM $M_L$
\Statex \textbf{Data:} Labeled $D_L = \{(x_l, y_l)\}$, Unlabeled $D_U = \{x_u\}$

\vspace{0.2cm}
\State \textbf{for} iteration $t = 1, \dots, T$ \textbf{do}

\State \quad \textcolor{gray}{\textit{// ASR inference}}
\State \quad $D'_L \leftarrow \{ (x_l, y_l, M_A(x_l)) \mid (x_l, y_l) \in D_L \}$
\State \quad $D'_U \leftarrow \{ (x_u, M_A(x_u)) \mid x_u \in D_U \}$

\State \quad \textcolor{gray}{\textit{// LLM training}}
\State \quad Train $M_L$ on $D'_L$

\State \quad \textcolor{gray}{\textit{// LLM inference}}
\State \quad $D''_U \leftarrow \{ (x_u, y'_u, M_L(x_u, \mathcal{T}(y'_u))) \mid (x_u, y'_u) \in D'_U \}$

\State \quad \textcolor{gray}{\textit{// Correction filtering (optional)}}
\State \quad $D^{f}_U \leftarrow \{ (x_u, y''_u) \mid (x_u, y'_u, y''_u) \in D''_U, \text{filter}(y'_u, y''_u) \}$

\State \quad \textcolor{gray}{\textit{// ASR training}}
\State \quad Train $M_A$ on $D_L \cup D^{f}_U$

\State \textbf{end for}
\State \textbf{return} $M_A$ and $M_L$

\end{algorithmic}
\end{algorithm}

In the initial \textbf{ASR inference} stage, the current ASR model $M_A$ generates hypotheses for both the labeled dataset $D_L = \{(x_l, y_l)\}$ and the unlabeled dataset $D_U = \{x_u\}$. For the labeled data, this yields triplets $(x_l, y_l, y'_l)$, where $y'_l = M_A(x_l)$, which serve as training examples for the corrector. Simultaneously, initial noisy pseudo-labels $y'_u = M_A(x_u)$ are generated for the unlabeled data to act as input for correction. Subsequently, in the \textbf{LLM training} stage, the Audio LLM $M_L$ is fine-tuned to perform ASR error correction. Its input concatenates encoded audio embeddings followed by the tokenized hypothesis prompt $\mathcal{T}(y'_l)$, so the model conditions on acoustic evidence before the text. It learns to map these multimodal inputs to the ground truth $y_l$ by minimizing the supervised fine-tuning loss $\mathcal{L}_{LLM} = -\sum \log P(y_l \mid x_l, \mathcal{T}(y'_l); \theta_L)$, where $\mathcal{T}(\cdot)$ represents a prompt template wrapping the hypothesis. This step effectively conditions the LLM to identify and repair specific error patterns produced by the ASR model, leveraging both acoustic evidence and linguistic priors. The process then advances to the \textbf{LLM inference} stage, where the fine-tuned $M_L$ rectifies the noisy pseudo-labels $y'_u$ from the unlabeled set, producing refined transcriptions $y''_u = M_L(x_u, \mathcal{T}(y'_u))$. To safeguard quality, an optional filtering mechanism removes corrections exhibiting signs of hallucination or excessive deviation, resulting in a high-fidelity subset $D^{f}_U$. Finally, in the \textbf{ASR training} stage, $M_A$ is fine-tuned on a combined corpus of the original labeled data and the filtered, corrected pseudo-labels. The objective minimizes $\mathcal{L}_{ASR}$, defined as a sum of the supervised loss on $D_L$ and the semi-supervised loss on $D^{f}_U$. This updated $M_A$ is then carried over to the next cycle, closing the loop. The iterative process terminates when a predefined maximum iteration count is reached or when validation performance saturates.

To further safeguard the training process against occasional hallucinations generated by the LLM, we explore two filtering strategies. The first is a heuristic rule-based mechanism that imposes constraints on character deviation and length consistency relative to the original ASR hypothesis, rejecting samples that exhibit significant divergence or repetition loops. The second is a model-based approach where a pre-trained text encoder is fine-tuned as a binary classifier to predict whether a given correction yields a lower word error rate (WER) than the initial hypothesis. By integrating this audio-aware correction, optionally augmented by robust filtering, ReHear mitigates the noise accumulation typical of standard pseudo-labeling and progressively leverages unlabeled data for sustained gains.

%%%%%%%%%%%%%%%%%%%%%%%%%%%%%%%%%%%%%%%%
\section{Experiments}

In this section, we outline the datasets, experimental setup, main results, and ablation studies.

%%%%%%%%%%%%%%%%%%%%%%%%%%%%%%%%%%%%%%%%
\subsection{Datasets}

We evaluate on four corpora: Earnings-21~\cite{del2021earnings}, Earnings-22~\cite{del2022earnings}, SPGISpeech~\cite{oneill2021spgispeech}, and the AMI Meeting Corpus~\cite{carletta2007unleashing,renals2007recognition}. Earnings-21 provides approximately $39$ hours of earnings calls across nine financial sectors, serving as a benchmark for entity-dense speech and named entity recognition. Building on this, Earnings-22 provides $119$ hours of recordings from global companies, specifically designed to test system robustness against diverse international accents. SPGISpeech is a large-scale corpus originally containing $5,000$ hours of formatted earnings calls; to facilitate efficient experimental iteration, we use a representative sampled subset. Finally, AMI comprises $100$ hours of multi-party meeting recordings with unscripted dynamics and frequent disfluencies, assessing generalization to spontaneous conversational speech. All corpora consist of English speech across diverse domains and accents. Table~\ref{tab:datasets} summarizes partition splits and statistics.

\begin{table}[h]
\centering
\caption{Dataset statistics with total duration per partition split and average segment duration.}
\label{tab:datasets}
\resizebox{\linewidth}{!}{%
\begin{tabular}{lrrrr}
\toprule
\textbf{Split} & \textbf{Earnings-21} & \textbf{Earnings-22} & \textbf{SPGISpeech} & \textbf{AMI} \\
\midrule
Labeled Train & 10.76h & 21.81h & 19.46h & 16.37h \\
Labeled Val & 3.07h & 9.87h & 9.24h & 8.94h \\
Labeled Test  & 4.93h & 12.41h & 11.43h & 8.68h \\
Unlabeled & 18.86h & 63.67h & 58.43h & 61.49h \\
\midrule
\textit{Total} & 37.63h & 107.75h & 98.55h & 95.48h \\
\textit{Avg. Duration} & 24.76s & 24.13s & 9.16s & 2.56s \\
\bottomrule
\end{tabular}
}
\end{table}

%%%%%%%%%%%%%%%%%%%%%%%%%%%%%%%%%%%%%%%%
\subsection{Data Preparation}

For the long-form Earnings-21 and Earnings-22 datasets, we standardize raw recordings to $16$kHz and normalize transcripts by mapping non-speech annotations (e.g., laughter) to unified special tags. To generate training-ready samples, we employ a segmentation pipeline using a pre-trained Wav2Vec 2.0 model~\cite{baevski2020wav2vec} coupled with CTC segmentation~\cite{graves2006connectionist,kurzinger2020ctc}. We utilize sliding window inference for acoustic probabilities and align them with text using a backtracking window of $8,000$ to $100,000$ frames. Adjacent segments with silence gaps under $2$s are merged, targeting a maximum chunk duration of $30$s after $0.1$s boundary padding. For the pre-segmented SPGISpeech and AMI corpora, we bypass alignment. The SPGISpeech dataset (small version) is downsampled to create a representative subset: we sample $10\%$ of training as labeled and $30\%$ as unlabeled, while retaining $10\%$ of both validation and test splits. For AMI independent headset microphones (IHM), we perform session-level partitioning, selecting $20\%$ of training sessions as labeled and the remainder as unlabeled. We discard segments shorter than $0.05$s, longer than $30$s, or with empty transcripts, and partition at the source-file or meeting level to prevent speaker or context leakage.

%%%%%%%%%%%%%%%%%%%%%%%%%%%%%%%%%%%%%%%%
\subsection{Prompt Templates}

To adapt the LLM for the ASR correction task and ensure the output meets pseudo-labeling requirements (i.e., transcription only), we design specific instruction templates. For the primary audio-aware correction, we instruct the model to ground its output in acoustic features using the following template: \textit{``Correct the ASR hypothesis based on the provided audio. Transcribe the speech exactly as spoken. Output strictly the corrected text without any explanations or fillers. Hypothesis: $<$hypothesis$>$''}, where \textit{$<$hypothesis$>$} is a placeholder for the ASR hypothesis. For the text-only variant where audio is unavailable, we modify this template by replacing the first two sentences with linguistic constraints: \textit{``Correct the ASR hypothesis by fixing typos and misspellings. Preserve the original style and do not paraphrase''}. The remaining instructions are kept identical to ensure the model strictly avoids generating extraneous conversational content.

%%%%%%%%%%%%%%%%%%%%%%%%%%%%%%%%%%%%%%%%
\subsection{Experimental Setup}

We utilize Whisper-Large-v3~\cite{radford2023robust} as the base ASR model, Voxtral-Mini-3B-2507~\cite{liu2025voxtral} as the Audio LLM corrector, and DeBERTa-v3-Base~\cite{he2021deberta} for model-based filtering. To ensure parameter efficiency, we employ Low-Rank Adaptation (LoRA)~\cite{hu2022lora} with rank $16$, scaling $32$, and dropout $0.05$. Both ASR and LLM backbones utilize 4-bit quantization (QLoRA)~\cite{dettmers2023qlora} with adapters applied to the ASR decoder and LLM language model, while the filter uses standard LoRA on its encoder. Training runs for $5$ epochs with a global batch size of $128$ and a cosine scheduler ($10\%$ warmup), using peak learning rates of $5 \times 10^{-4}$ for ASR, $10^{-4}$ for the LLM, and $10^{-3}$ for the model filter. Across iterations, the LoRA adapters are carried over rather than reinitialized, with learning rate and epoch count held constant. During inference, we employ greedy decoding for ASR and beam search (width $5$) for the LLM. For rule-based filtering, we discard corrections if the character error rate (CER) exceeds $0.15$, the length ratio falls outside $[0.95, 1.15]$, the unique token ratio is below $0.40$, or if digit mismatches exceed $2$. We evaluate ReHear against iterative supervised learning (ISL) and two variants of iterative pseudo-labeling (IPL)~\cite{xu2020iterative,likhomanenko2021slimipl}: a standard naive IPL utilizing raw ASR hypotheses, and a robust rule-filtered IPL. For the latter, we apply strict criteria to retain only high-quality pseudo-labels: confidence at least $0.95$, speaking rate within $[2.0, 5.0]$ words/sec, and compression ratio at least $0.5$. For each configuration, we conduct $5$ independent runs with $T=3$ outer iterations, selecting the checkpoint with the lowest validation WER. All reported WER metrics are computed after standard text normalization, which includes lowercasing, removing punctuation, and stripping common filler words. We implement our framework in PyTorch~\cite{paszke2019pytorch} with the HuggingFace Transformers library~\cite{wolf2020transformers}. All experiments were conducted on $8$ NVIDIA A100 GPUs.

%%%%%%%%%%%%%%%%%%%%%%%%%%%%%%%%%%%%%%%%
\subsection{Experimental Results}

\begin{table*}[!t]
\centering
\caption{WER (\%) mean and standard deviation (subscript); \textbf{bold} and \underline{underline} mark the best and second-best results.}
\label{tab:main-results}
\footnotesize
\begin{tabular}{lcccccccc}
\toprule
\multirow{2.5}{*}{\textbf{Method}} & \multicolumn{2}{c}{\textbf{Earnings-21}} & \multicolumn{2}{c}{\textbf{Earnings-22}} & \multicolumn{2}{c}{\textbf{SPGISpeech}} & \multicolumn{2}{c}{\textbf{AMI (IHM)}} \\
\cmidrule(lr){2-3} \cmidrule(lr){4-5} \cmidrule(lr){6-7} \cmidrule(lr){8-9}
& \textbf{Test} & \textbf{Unlabeled} & \textbf{Test} & \textbf{Unlabeled} & \textbf{Test} & \textbf{Unlabeled} & \textbf{Test} & \textbf{Unlabeled} \\
\midrule
ISL
& 6.76\sd{0.17} & \phantom{0}8.57\sd{0.19}
& 12.54\sd{0.11} & 11.43\sd{0.09}
& 2.41\sd{0.01} & 2.42\sd{0.02}
& \phantom{0}9.66\sd{0.24} & 10.26\sd{0.29} \\
IPL
& 8.02\sd{0.18} & 10.69\sd{0.10}
& 14.02\sd{0.17} & 13.30\sd{0.08}
& 2.83\sd{0.05} & 2.98\sd{0.08}
& 16.52\sd{0.26} & 19.31\sd{0.13} \\
IPL + Rule & 7.59\sd{0.25} & 10.30\sd{0.17} & 13.64\sd{0.12} & 13.10\sd{0.08} & 2.57\sd{0.03} & 2.59\sd{0.04} & 11.00\sd{2.05} & 12.35\sd{1.86} \\
ReHear
& \textbf{6.34}\sd{0.26} & \phantom{0}\underline{7.87}\sd{0.18}
& \underline{12.01}\sd{0.08} & \underline{10.63}\sd{0.11}
& \textbf{2.36}\sd{0.01} & \textbf{2.23}\sd{0.03}
& \phantom{0}\underline{9.45}\sd{0.23} & 10.30\sd{0.15} \\
ReHear + Rule
& \underline{6.35}\sd{0.13} & \phantom{0}\textbf{7.79}\sd{0.08}
& \textbf{11.82}\sd{0.16} & \textbf{10.63}\sd{0.16}
& \underline{2.37}\sd{0.04} & \underline{2.28}\sd{0.04}
& \phantom{0}\textbf{9.35}\sd{0.24} & \textbf{10.09}\sd{0.22} \\
ReHear + Model
& 6.55\sd{0.19} & \phantom{0}7.94\sd{0.15}
& 12.05\sd{0.09} & 10.78\sd{0.09}
& 2.39\sd{0.04} & 2.31\sd{0.04}
& \phantom{0}9.74\sd{0.20} & \underline{10.13}\sd{0.29} \\
\bottomrule
\end{tabular}
\end{table*}

Table~\ref{tab:main-results} reports WER on two splits: the held-out labeled test split, which measures generalization, and the unlabeled pool targeted for adaptation, whose references are used only to score WER while adaptation relies on pseudo-labels. Traditional iterative pseudo-labeling (IPL) often struggles to improve upon the supervised baseline (ISL), particularly on challenging datasets like Earnings and AMI where confirmation bias leads to performance degradation. Even when incorporating strict rule-based filtering (IPL + Rule) to reject confident errors, the method fails to consistently outperform the supervised baseline. In contrast, ReHear largely overcomes this issue: by leveraging audio-aware LLM corrections to generate high-fidelity targets, it improves over both ISL and IPL on nearly all test and unlabeled splits, and ReHear variants (standard and rule-filtered) attain the best results across all splits. Detailed analysis reveals that the proposed framework is particularly effective in complex acoustic environments, such as the spontaneous financial narratives in Earnings-21/22 and the multi-speaker meeting scenarios in AMI, achieving substantial WER reductions on both labeled and unlabeled sets. Even on SPGISpeech, where the baseline is already accurate, our method yields further gains. Regarding filtering strategies, rule-based constraints (ReHear + Rule) yield the highest stability, while the learned model filter (ReHear + Model) also delivers competitive results, albeit slightly constrained by overfitting on limited labeled data. Notably, standard ReHear without additional filtering remains highly robust and frequently surpasses filtered IPL, demonstrating that the audio-aware corrector significantly reduces the dependency on complex post-hoc filtering mechanisms.

%%%%%%%%%%%%%%%%%%%%%%%%%%%%%%%%%%%%%%%%
\subsection{Ablation Studies}

To investigate the contribution of specific design choices, we conduct ablation studies on the Earnings-21 dataset, reporting all results using the standard ReHear framework without additional filtering mechanisms.

%%%%%%%%%%%%%%%%%%%%%%%%%%%%%%%%%%%%%%%%
\subsubsection{Modality and Position}

We analyze the contribution of acoustic features and their positional encoding within the input sequence. Table~\ref{tab:modality-results} shows that text-only input yields substantially higher WER, indicating that without acoustic grounding the LLM acts as a grammatical editor unable to fix phonetic errors. Incorporating audio produces large gains on both sets, with audio-prefix (audio + text) achieving the best performance. This suggests that conditioning on the source speech before the erroneous text improves error detection and correction.

\begin{table}[!h]
\centering
\caption{Ablation on input modality and audio positioning on Earnings-21 (WER \%).}
\label{tab:modality-results}
\footnotesize
\begin{tabular}{lcc}
\toprule
\textbf{Configuration} & \textbf{Test} & \textbf{Unlabeled} \\
\midrule
Text Only & 7.74\sd{0.11} & 10.46\sd{0.12} \\
Text + Audio & \underline{6.34}\sd{0.24} & \phantom{0}\underline{8.02}\sd{0.24} \\
Audio + Text & \textbf{6.34}\sd{0.26} & \phantom{0}\textbf{7.87}\sd{0.18} \\
\bottomrule
\end{tabular}
\end{table}

%%%%%%%%%%%%%%%%%%%%%%%%%%%%%%%%%%%%%%%%
\subsubsection{Decoding Strategies}

We evaluate the impact of different decoding strategies on the quality of Audio LLM-generated corrections. Table~\ref{tab:decoding-results} shows that deterministic beam search yields substantially lower WER than greedy decoding or stochastic sampling ($T=0.7$), whose randomness harms an error-correction task where fidelity to the source audio matters more than generation diversity. In contrast, beam search facilitates a more effective exploration of the hypothesis space. Beam width $N=3$ is best on labeled test and $N=5$ best on unlabeled; we adopt beam search for robust, acoustically grounded pseudo-labels.

\begin{table}[!h]
\centering
\caption{Ablation on LLM decoding strategies (greedy, beam search, and stochastic sampling) on Earnings-21 (WER \%).}
\label{tab:decoding-results}
\footnotesize
\begin{tabular}{llcc}
\toprule
\textbf{Strategy} & \textbf{Configuration} & \textbf{Test} & \textbf{Unlabeled} \\
\midrule
Greedy & $N=1$ & 6.59\sd{0.36} & 8.67\sd{0.34} \\
Beam & $N=3$ & \textbf{6.21}\sd{0.14} & \underline{7.89}\sd{0.19} \\
Beam & $N=5$ & \underline{6.34}\sd{0.26} & \textbf{7.87}\sd{0.18} \\
Sampling & $T=0.7$ & 6.54\sd{0.22} & 8.72\sd{0.23} \\
\bottomrule
\end{tabular}
\end{table}

%%%%%%%%%%%%%%%%%%%%%%%%%%%%%%%%%%%%%%%%
\subsubsection{Iterative Dynamics}

We investigate how the self-training loop depth affects performance on Earnings-21, comparing training with and without a decay strategy that halves the learning rate and decrements epochs each iteration to reduce overfitting and noise accumulation. Table~\ref{tab:iteration-results} reports WER by iteration, where iteration $0$ is the zero-shot base model before any fine-tuning and each subsequent iteration adds one round of audio-aware correction and student fine-tuning. Gains realize rapidly and saturate around iteration two; without decay, WER degrades beyond that, while decay stabilizes training over longer loops. We set the limit to $3$ iterations to balance efficiency and stability, since extending the loop yields diminishing returns.

\begin{table}[!h]
\centering
\caption{Ablation on iterative dynamics on Earnings-21 with and without training decay (WER \%).}
\label{tab:iteration-results}
\resizebox{\linewidth}{!}{%
\begin{tabular}{ccccc}
\toprule
\multirow{2.5}{*}{\textbf{Iteration}} & \multicolumn{2}{c}{\textbf{Without Decay}} & \multicolumn{2}{c}{\textbf{With Decay}} \\
\cmidrule(lr){2-3} \cmidrule(lr){4-5}
& \textbf{Test} & \textbf{Unlabeled} & \textbf{Test} & \textbf{Unlabeled} \\
\midrule
0 & 8.98\sd{0.00} & 10.99\sd{0.00} & 8.98\sd{0.00} & 10.99\sd{0.00} \\
1 & \underline{6.48}\sd{0.20} & \phantom{0}\underline{8.06}\sd{0.12} & \textbf{6.30}\sd{0.18} & \phantom{0}8.06\sd{0.29} \\
2 & \textbf{6.35}\sd{0.10} & \phantom{0}\textbf{7.92}\sd{0.15} & 6.43\sd{0.22} & \phantom{0}\textbf{7.94}\sd{0.13} \\
3 & 6.51\sd{0.21} & \phantom{0}8.07\sd{0.17} & \underline{6.42}\sd{0.16} & \phantom{0}\underline{7.99}\sd{0.12} \\
4 & 6.75\sd{0.26} & \phantom{0}8.26\sd{0.29} & 6.48\sd{0.16} & \phantom{0}8.05\sd{0.19} \\
5 & 6.77\sd{0.08} & \phantom{0}8.34\sd{0.28} & 6.53\sd{0.14} & \phantom{0}8.06\sd{0.19} \\
\bottomrule
\end{tabular}
}
\end{table}

%%%%%%%%%%%%%%%%%%%%%%%%%%%%%%%%%%%%%%%%
\section{Conclusion}

In this paper, we introduced a semi-supervised framework that employs instruction-tuned audio-aware large language models for speech recognition correction and iterative pseudo-labeling. By grounding the correction process directly in acoustic features, our method generates high-quality supervision signals that surpass those derived from filtered pseudo-labels and text-only correction. Empirical evaluations across multiple datasets confirm that this framework improves over supervised benchmarks and traditional pseudo-labeling techniques on nearly all splits, most consistently with correction filtering. Our analysis further demonstrates that incorporating audio context is the decisive factor in reducing recognition errors, offering a data-efficient pathway for enhancing system performance using unlabeled speech.

%%%%%%%%%%%%%%%%%%%%%%%%%%%%%%%%%%%%%%%%
\bibliographystyle{IEEEtran}
\bibliography{refs}
%%%%%%%%%%%%%%%%%%%%%%%%%%%%%%%%%%%%%%%%
\end{document}